\documentclass{article}
\usepackage{ijcai21}

\usepackage{times}

\usepackage{soul}
\usepackage{url}
\usepackage[hidelinks]{hyperref}
\usepackage[utf8]{inputenc}
\usepackage[small]{caption}
\usepackage{graphicx}
\usepackage[table,xcdraw]{xcolor}
\usepackage{amsfonts,amsmath,amssymb,amsthm}
\usepackage{booktabs}
\usepackage{lipsum} 
\usepackage{csquotes}
\usepackage{overpic}
\usepackage{wrapfig}
\usepackage{multirow}
\DeclareMathOperator*{\argmax}{arg\,max}

\theoremstyle{definition}

\theoremstyle{definition}
\newtheorem*{roadmap}{Roadmap}

\theoremstyle{definition}

\title{Where the Action is: Let's make Reinforcement Learning for Stochastic Dynamic Vehicle Routing Problems work!}

\author{
Florentin D Hildebrandt$^1$\footnote{Contact Author}\and
Barrett Thomas$^2$\and
Marlin W Ulmer$^{1}$\\
\affiliations
$^1$TU Braunschweig\\
$^2$University of Iowa\\
\emails
f.hildebrandt@tu-braunschweig.de,
barrett-thomas@uiowa.edu,
m.ulmer@tu-braunschweig.de
}

\begin{document}

\maketitle

\begin{abstract}
There has been a paradigm-shift in urban logistic services in the last years; demand for real-time, instant mobility and delivery services grows. This poses new challenges to logistic service providers as the underlying stochastic dynamic vehicle routing problems (SDVRPs) require anticipatory real-time routing actions. \emph{Searching} the combinatorial action space for efficient routing actions is by itself a complex task of mixed-integer programming (MIP) well-known by the operations research community. This complexity is now multiplied by the challenge of \emph{evaluating} such actions with respect to their effectiveness given future dynamism and uncertainty, a potentially ideal case for reinforcement learning (RL) well-known by the computer science community. For solving SDVRPs, joint work of both communities is needed, but as we show, essentially non-existing. Both communities focus on their individual strengths leaving potential for improvement. Our survey paper highlights this potential in research originating from both communities. We point out current obstacles in SDVRPs and guide towards joint approaches to overcome them.

%To address this novel combination, hybrid approaches are needed because SDVRP is a child of two worlds: Anticipatory real-time decision making suits RL methods while combinatorial state and action space call for MIP approaches. This makes SDVRP a great opportunity to foster joint developments in reinforcement learning and operations research. 

%\color{red}
%\begin{itemize}
%    \item SDVRPs become important
%    \item well-established static-VRP-MIP-tools are only part of the solution
%    \item real-time anticipatory decision making $\rightarrow$ ideal playground for RL, COP-challenge
%    \item VRP comes from the decision space, RL comes from the value function, meeting point is needed.
%    %\item RL Community looks at static OR problems where runtime is secondary and MIP solver reign supreme
%    %\item But there is a whole class of OR problems that come naturally as MDP and require real-time anticipatory decision making $\rightarrow$ ideal playground for RL
%    \item survey paper highlights the gaps between OR/MIP community and RL community
%    \item we show what has been done and what could be done
%\end{itemize}
%\color{black}
\end{abstract}

\section{Introduction}
The world became more interconnected with the fall of the iron curtain in 1991. The resulting global market community posed logistical and industrial challenges that required novel decision making and planning tools. At this time, computing became much cheaper, faster, and had greater memory storage, starting the golden age of mixed integer programming (MIP). A new generation of exact solvers, building on decades of theoretical work and advances in computer technologies, were suddenly capable of solving combinatorial optimization problems (COP) with millions of variables and thousands of constraints ( \cite{bixby2012brief}). This combination of logistical challenges and technical advancements allowed to put theory into practice and logistics optimization made a great leap forward. In particular planning problems like NP-hard vehicle routing problems (VRPs) were perfect applications for exact solvers because solution quality was pivotal, runtime secondary, and \emph{static} environments were assumed.

Today, the world is more \emph{dynamic} – and so are the optimization problems! Global interconnectedness, urbanization, ubiquitous information streams, and increased service-orientation raise the need for anticipatory real-time decision making (see \cite{archetti2021recent}).
A striking example are again logistic service providers: Service promises, like same-day or restaurant meal delivery, dial-a-ride, and emergency repair, force logistic service providers to anticipate future demand, adjust to real-time traffic information, or even incorporate unknown crowdsourced drivers to fulfill customer expectations. Provably optimal, but \emph{static} routing solutions are no longer sufficient to overcome the challenges of such services, \emph{dynamic} (re-)routing is required.\par

A new problem class called stochastic dynamic vehicle routing problem (SDVRP) closes the gap between VRP modelling and \emph{dynamic} real-world routing problems by introducing an additional time axis. The planning objective is no longer limited to \emph{searching} larger and larger routing action spaces but also focuses on \emph{evaluating} actions with respect to future uncertainty.
%SDVRPs are characterized by repeated actions of re-planning (dynamism) along this time axis to address uncertainty in the exogenous process (stochasticity). 
SDVRPs fall into the domain of dynamic combinatorial optimization, a sub-field of OR, that addresses sequential decision problems with future uncertainty. The goal is to derive decision policies instead of static solutions assigning a routing action to each state of the system. Such a routing action is usually represented as a sequence of tentative stops for each vehicle that satisfies the specific problem-constraints such as time-windows, deadlines, pickup before delivery, or vehicle capacity. In theory, reinforcement learning (RL) appears to be an ideal solution method for dynamic combinatorial optimization and SDVRPs because
\begin{enumerate}
    \item[(a)] dynamic combinatorial optimization problems can naturally be modeled as Markov decision processes (MDP) and
    \item[(b)] deep neural networks (DNN) are able to serve as fast and complex decision policies.
\end{enumerate}

However, in SDVRP-work of the OR-community, the majority of available RL-techniques remains nearly untouched. Instead, alternative OR-strategies are applied, often based on the community's well-known expertise on MIPs and focused more on searching the space of routing actions in a state and less on their evaluation. Further, while there is an increasing amount of RL-work on SDVRPs in the computer science (CS) community, the considered problems are usually comparably simple in action and state space; thus, searching the action space is relatively easy and the community can focus on the evaluation of actions instead. While each community plays to their strengths, there is room for improvement, and, as we show in this review, room for collaboration.\par

\begin{roadmap}
In this survey, we present the SDVRP class and its challenges to the broader scientific community. We analyze the literature on reinforcement learning for SDVRP from the OR and CS perspectives. We identify differences between the world of mixed integer programming of OR and reinforcement learning of CS. Exact MIP solvers search the action space effectively; RL methods evaluate actions well with respect to the future. We highlight that existing solution methods draw from either side and are generally unable to do both. We provide guidance on how future research may bridge the gap. 
\end{roadmap}

\section{Stochastic Dynamic Vehicle Routing}\label{sec:sdvrp}
%\lipsum[7-9]
In this section, we give an overview on SDVRPs. We first briefly recall static VRPs and then extend them to SDVRPs. Finally, we provide a short history of SDVRP-solution methods.

\subsection{Static Vehicle Routing Problems}
In static VRPs, all information is known at time of decision making. There are many applications that induce static VRPs, among others are parcel or grocery delivery, pickup and delivery of (known) passengers or goods, or routing of a technician crew to repair or maintain equipment. In static VRPs, actions are usually made about the \emph{routing}, i.e. assignment of customers to vehicles and the sequencing of the vehicle stops. Routing actions must always satisfy constraints like the well-known \enquote{subtour elimination constraints} and sometimes must satisfy application specific constraints like capacity constraints, time-window constraints, and pickup-before-delivery constraints. A common objective is to maximize service (e.g., the number of customers served) or to minimize cost (e.g., travel time or distance).

%\textcolor{red}{wollen/können wir das im verlauf des papers mehr einbeziehen? PS Ich könnte mir eine SL-methode vorstellen, der man A, b, und ein x gibt, und die sagt, wie wahrscheinlich es ist, dass das feasible ist. Sowas würde ich in den Outlook packen.}
In essence, we can model static VRPs as a MIP of the form:

\begin{equation}\label{eq:static}
\begin{array}{ll@{}ll}
\text{minimize}  & C &x  &\\
\text{subject to}& A   &x \leq b, \\
                 &                                                &x \in \{0,1\}^n &
\end{array}
\end{equation}

Solutions are modeled as a vector $x$ of binary variables\footnote{We note that dependent on specific problems and the modeling style, parts of the solution variables may be integer or of real values.}.
The objective is a linear function given by a cost vector $C$ and the solution $x$ taken. The solutions are required to satisfy a set of linear constraints, summarized in matrix $A$ and limited by vector $b$.

The number of solution variables regularly explodes when static VRPs increase in size. Additionally, constraints are often complex and even finding a feasible solution is challenging. 
%\textcolor{red}{lass uns hier und für die anderen mal ein paar overview/erste paper nennen, um die leute anzuziehen}
The OR-community has developed several strong solution mechanisms to tackle these NP-hard MIPs, including among others, Benders-decomposition, Column generation, Branch-cut-and-price, or Dynamic discretization discovery (see, for example, \cite{toth2014vehicle}, \cite{braekers2016vehicle}, and \cite{boland2017continuous}). %(\textcolor{red}{Barry, we need your help here. Is this correct and did we miss something? Also, we should cite people we want to point towards our research}). 
All methodologies exploit structures in action variables and constraints to systematically search the vast solution space.

\subsection{Stochasticity and Dynamism}
With the paradigm shift in the logistic business models towards instant gratification and real-time fulfillment, the nature of many routing problems has become increasingly stochastic and dynamic. Routing actions are made under incomplete information and the problem requires frequent updates or entire re-planning. Examples for SDVRPs are courier or dial-a-ride services (e.g., \cite{ghiani2009anticipatory},\cite{agatz2012optimization}), restaurant meal or instant delivery (e.g., \cite{yildiz2019provably}, \cite{voccia2019same}), as well as emergency repair or healthcare services (e.g., \cite{schmid2012solving}). Uncertainties are manifold and manifest in the customers (e.g., demand volumes, new requests, choices, \cite{bent2004scenario}, \cite{ulmer2020dynamic}), in the environment (e.g., travel times or parking, \cite{schilde2014integrating}), and increasingly also in the fleet itself (e.g., crowdsourced drivers and their behavior, \cite{arslan2019crowdsourced}, \cite{ulmer2020workforce}).

The result are stochastic dynamic decision problems that are generally modelled via Markov decision process (MDP, \cite{ulmer2020modeling}). An MDP models a problem as a sequence of states $S_k$, actions $x_k$, reward function $R(S_k,x_k)$ and value function $V(S_k,x_k)$, revelations of stochastic information $W_{t+k}$, and a transition function $S^M(S_k,x_k,W_{k+1})$ towards the next state $S_{k+1}$ over a set of time steps $t_1,\dots,t_K$ in the problem time horizon. For SDVRPs, the MDPs typically take the following form:\\

 \noindent \textbf{States $S_k$:} A state comprises all information needed to select an action, thus, information about the customers (locations, time-windows, status, etc.), the fleet (position, tentative routing, status, etc.), and the environment (traffic  or parking situation, etc.). In essence, a state of the MDP shows similarities with the MIP presented in Equation~(\ref{eq:static}) with state-dependent matrix $A(S_k)$ and limiting vector $b(S_k)$.\\
 
\noindent \textbf{Actions $x_k$:} Actions update the tentative routing of the fleet. Thus, the action space is equivalent to the solution space of the MIP presented in Equation~(\ref{eq:static}). However, routing actions are not permanent but rather tentative and can be partially altered in later states.\\ %\textcolor{red}{Brauchen wir glaube ich doch nicht mehr.}The (deterministic) combination of state $s_k$ and decision $x_k$ is also called a post-decision state $s^x_k$ representing the state updated by the decision made.\\

 \noindent \textbf{Reward function $R(S_k,x_k)$:} The reward function evaluates the immediate impact of an action on the objective value, e.g., routing cost or increase in served customers. It is similar but not necessarily identical to the cost-function of the MIP in Equation~(\ref{eq:static}).\\
 
 \noindent \textbf{Value function $V(S_k,x_k)$:} The value function depicts the maximum (or minimum) expected value that can be achieved in the future when taking action $x_k$ in state $S_k$. Having access to the value function assumes the possibility of optimal decision making for the rest of the horizon and of capturing all potential changes in information. It is not considered in the MIP in Equation~(\ref{eq:static}).\\ %(A theoretical value function would always map to zero in that static case since there are no further actions and rewards.).\\
 
 \noindent \textbf{Stochastic information $W_{k+1}$:} After an action $x_k$ in $S_k$ is taken, a realization of the stochastic information $W_{k+1}$ is revealed. For SDVRPs, such information may change information about customers, fleet resources, or the environment the fleet operates in. With respect to the MIP in Equation~(\ref{eq:static}), the information may impact the constraint matrix $A$, or the limiting vector $b$. It may also impact the set of action variables.\\
 
  \noindent \textbf{Transition function $S^M(S_k,x_k,W_{k+1})$:} Based on state $S_k$, action $x_k$, and stochastic information $W_{k+1}$, the transition function leads to a new state $S_{k+1}$ by adapting constraint matrix, limiting vector, and action variables.\\

 \noindent \textbf{Policies:} A solution of the MDP is a policy $X^\pi$ that assigns an action $x^\pi_k$ to every state $S_k$. The objective of the MDP is to find an optimal policy $X^*$ that maximizes the expected reward over the entire time horizon. An optimal policy satisfies the Bellman Equation, maximizing the immediate reward plus the value of an action in a state:
 
\begin{equation}
    x^*_k= \argmax\limits_{x_k: A(S_k)x_k\leq b(S_k)}\{R(S_k,x_k)+V(S_k,x_k)\}.
\end{equation}

\noindent When solving the SDVRP, two components are particularly interesting; the action space and the value function. As previously discussed, searching the action space is equivalent to finding a solution for the MIP in Equation~(\ref{eq:static}). However, in stark contrast to the static MIP, the values of the actions are not known (except in the very rare case when it is possible to solve the remaining tree of the MDP exactly, e.g., via recursion). Thus, solving SDVRPs combines two substantial challenges, \emph{searching} an action in a complex, NP-hard mixed-integer program and \emph{evaluating} each action with respect to future information changes and actions. %These unique challenges have fostered the growth of modelling and algorithmic approaches, which we discuss in the next section.

\subsection{A Short History of SDVRP-Methods}

In this section, we briefly give an overview on the development of SDVRP-methods in the operations research community. For a detailed description and references, we refer to \cite{ulmer2020modeling}.
Early works on SDVRPs appeared around the turn of the millennium. Confronted with the new nature of stochasticy and dynamism, researchers turned to the well-known MIP-solution methods, thus ignoring future and uncertainty and treating a state of the MDP as an isolated instance on a rolling-horizon basis (e.g. \cite{gendreau1999parallel}). Since such \enquote{myopic} procedures often provided inflexible and eventually ineffective solutions, researchers looked for alternatives (\cite{powell2000value}). A first remedy was to apply on simple strategies mimicking practical decision making such as first-in-first-out, waiting, or threshold-strategies (e.g., \cite{thomas2007waiting}), see \enquote{policy function approximation} in \cite{powell2019unified}. Such methods generally ignore the vast number of potential actions as well as reward and value function. A similar idea is to manipulate the MIP to incentivize flexibility or avoid myopic decision making, for example, via safety buffers or penalty terms (see \enquote{cost function approximation} in \cite{powell2019unified}). This allowed exploitation of the powerful MIP-methodology while simultaneously achieving some anticipation of future changes in information and planning. With the increase in computational power and data availability, data-driven methods were developed to capture the uncertainty explicitly (e.g., \cite{bent2004scenario}). Such methods usually generated scenarios and solved either the MIPs augmented by the scenarios or the now deterministic and smaller multi-stage program (see \enquote{direct lookahead models} in \cite{powell2019unified}). While such methods are able to capture the future in detail, they are very runtime-demanding and tend to deviate from reality the longer the scenario horizon becomes (see \cite{voccia2019same}). To counter these insufficiencies, in the last years, first reinforcement learning methods were introduced as we will discuss in the next section.

\section{Solving SDVRPs with RL}\label{sec:solution}
\begin{wrapfigure}{L}{0.25\textwidth}
        \begin{overpic}[width=0.25\textwidth]{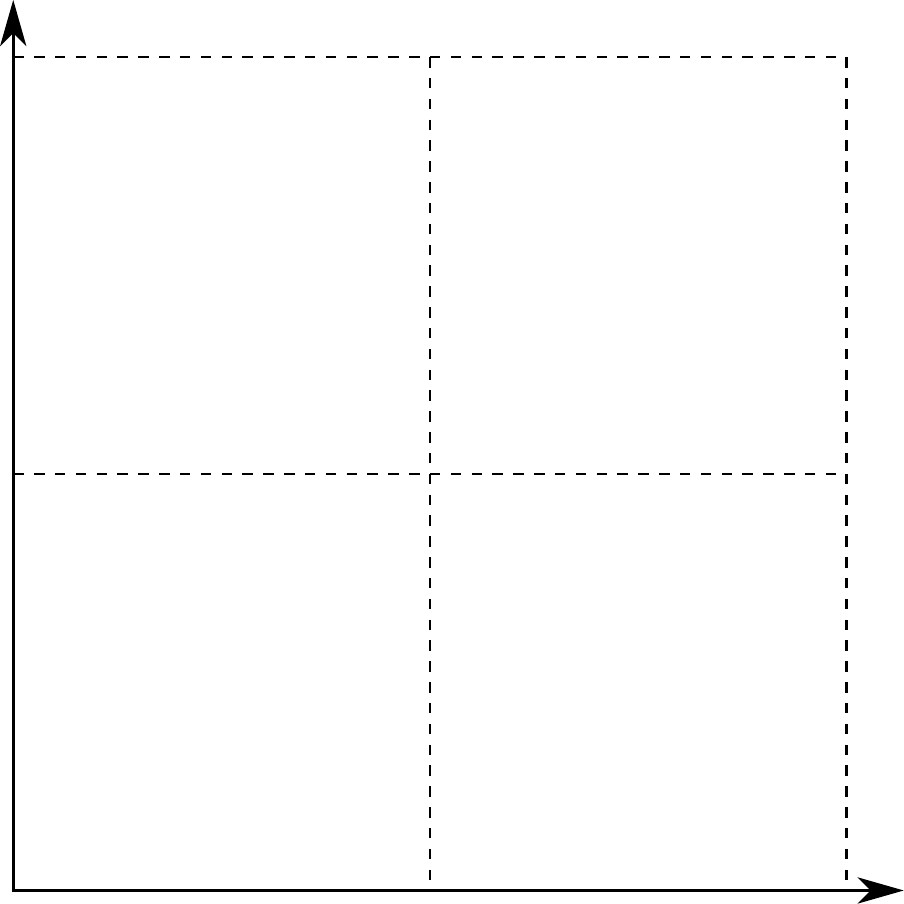}
        \put (15,65) {\huge CS}
        \put (20,3) {\tiny low}
        \put (65,3) {\tiny high}
        \put (60,20) {\huge OR}
        \put (-7,20) {\rotatebox{90}{State Action Detail}}
        \put (3,65) {\rotatebox{90}{\tiny fine}}
        \put (3,20) {\rotatebox{90}{\tiny coarse}}
        \put (20,-7) {Problem Complexity}
        \end{overpic}
    \hspace{1pt}
    \caption{Problem Complexity and State Action Detail of SDVRPs considered in CS and OR.}
    \label{fig:problem_dimensions}
\end{wrapfigure}
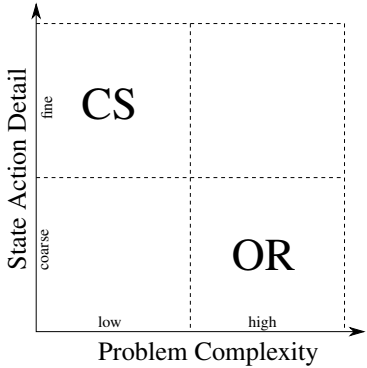

In reinforcement learning (RL), an agent learns by interacting with a partially observable environment. The agent aims to maximize a delayed reward either by parameterizing the state-dependent action distribution directly or by approximating state-action values. In most RL problems, the crux is to \emph{evaluate} actions in a given state. \emph{Searching} the action space is typically not necessary because most instantiations of RL require that we are able to enumerate an action space, e.g. in Q-learning. SDVRPs can have large, combinatorial actions spaces for which it might be impossible to enumerate the action space. The few existing works follow two streams of research originating from the operations research (OR) community and the computer science (CS) community. Even though there are some similarities in modeling and methodology, considerable differences remain. The streams can be differentiated by the \emph{complexity} of the SDVRPs and the level of \emph{detail} states and actions are considered in the solution approach, as we illustrate in Fig.~\ref{fig:problem_dimensions}. The x-axis denotes the problem complexity from low to high. Low-complexity SDVRPs resemble resource allocation problems while high-complexity SDVRPs require the construction of tentative routes to determine the feasibility of actions in a given state. The y-axis denotes the state action detail from coarse to fine. Many RL-methods restrict the action space and aggregate the state space to a small set of features. The level depends on the severity of state space aggregation and action space restriction.\par

% That is, how much of the available information is used in decision making and to what extent decision making relies on reinforcement learning.

As we will illustrate later in this section, RL-work of the OR-community is usually considering high problem complexity and limited state and action detail. OR-research on SDVRP is often motivated by specific practical applications. As such, they aim to model as many real-world aspects as possible. In particular, applications that are subject to time-window constraints (e.g. in service vehicle routing) or precedence constraints (e.g. in pickup-and-delivery problems) prove difficult because feasibility of assignment actions depend on the existence of tentative routes that fulfill aforementioned constraints. The interweaving of routing and assignment actions severely restricts solution approaches. %Employed RL methods are typically simple and constitute only a fragment of the solution, i.e., they are accompanied by established heuristics to reduce the action space and severe aggregation of the state space to a set of problem-specific features. \par
Employed RL methods are often simple and rely on enumeration of the action space. The required restriction of the action space is achieved by solving only parts of the problem by anticipatory RL methods, e.g., assignment actions, and other parts by hand-crafted heuristics, e.g., routing actions. Thus, RL methods are only a fragment of the solution method.\par

In contrast, CS-research on SDVRPs is often of low problem complexity and fine state and action detail. CS-work on SDVRP focuses primarily on analyzing the power and versatility of genuine RL methodology. As such, they tend to simplify problem complexity to allow the direct operation of their solution method on full action space and state space information. A common example are ride-sharing problems where pickup and delivery are merged to a single operation and vehicles are dynamically assigned to tiles in a discretized service area to serve demand, relocate, or re-charge. Actions determine only the next tile to visit for each vehicle and no tentative routes are planned. Thus, complex routing constraints, e.g. temporal commitments, cannot be integrated in such a modelling approach.\par

In the following, we analyze the gaps between works on SDVRP originating from OR and CS in detail. We summarize what has been done in each field and illustrate how a common effort could lead to the next step: Solving high-complexity SDVRPs using fine action state detail and sophisticated RL-methodology. 
%\textcolor{red}{müssen wir irgendwo nochmal search and evaluate rein packen? ggf ganz am Anfang der section, wenn wir RL beschreiben?}

\subsection{The OR Stream}
Many SDVRPs considered by OR are characterized by the need for explicitly constructing tentative routes to satisfy constraints. Planning with tentative routes introduces a further problem dimension that separates many SDVRPs from resource allocation problems. Assignment actions cannot be made without routing actions in these SDVRPs. This is an immense challenge.\par
%\textcolor{red}{gibts das problem? ansonsten würde ich einfach pickup and delivery with deadlines machen. da gibt es paper von Schilde 2014 in EJOR. Vielleicht auch lieber kürzen und mit vorherigem mergen, dann haben wir mehr platz.}
%In dynamic service vehicle routing, we must decide whether to offer a requesting customer a time-slot on the same day or serve the customer the next day. This entails checking if it is possible to assign the customer to a vehicle in our fleet such that there exists a tentative route that meets the customer's time slot without violating previous customers' time slots. Proposing a time slot, making an assignment, and planning a route in real-time while accounting for feasibility and future uncertainty seems to be an insurmountable task.
For example, in pickup-and-delivery problems with deadlines we must decide if we can serve a requesting customer. This entails checking if it is possible to assign the customer to a vehicle in our fleet, to pick the customer up and drop the customer off such that there exists a tentative route that meets the customer's deadline without violating previous customers' deadlines. As such, a service action cannot be made without consideration of routing actions.\par

 %(\textcolor{red}{Ich würde es mit den routes verbinden. Wir haben einen komplexen Action-Space den wir in real-time durchsuchen müssen. Deshalb restricition. Dann haben wir einen komplexen state space, da er routes enthält. Deshalb aggregation. Warum VFA? Habe keine Antwort. Ggf. weil Q-learning zweimal routes drin hat? - ich verstehe die Idee, aber um platz zu sparen, würde ich einfach nur einen paragraphen draus machen. Leute können 6 seiten lesen}):
%\begin{description}
%\item[Ingredients:] An action space restriction, a state space aggregation, and a value function approximation.
%\item[Instructions:] First, we greatly restrict the action space by solving only parts by RL and everything else by established heuristics. Second, we aggregate the state space to a small set of relevant features. Third, we learn the value of actions in a given, reduced state. Fourth, we search the restricted action space with the help of the approximated action values to derive actions.
%\end{description}
\emph{Searching} and \emph{evaluating} the action space is difficult for such complex problems.
Solution approaches reduce to a simple recipe: restrict the action space and aggregate the state space. Methods restrict the complex action space by solving only parts by RL and everything else by established heuristics, for example by learning the service offering and using an insertion heuristic to integrate the additional stops in the routes (see \cite{chen2019deep}). Furthermore, methods have to tackle a high-dimensional state space that includes the routes of all vehicles. Thus, they aggregate this state space to a few meaningful features, often associated with route details such as arrival times or slack (see \cite{ulmer2018budgeting}). The method then approximate the value of state-action tuples in the aggregated state space and restricted action space.\par

Such an heuristic approach enables anticipatory decision making in problems with complex route-constraints but might be too coarse. A major goal of future work is relaxing required simplifications towards finer state-action details. In essence, the OR-community aims to \emph{evaluate} state-action values more precisely and \emph{search} the action space more effectively. We guide future research by first analyzing past works and future opportunities in the \emph{search}-dimension before we investigate the \emph{evaluation}-dimension.\par

\subsubsection{Search.}
The majority of OR work on SDVRP that uses RL is focused on the challenges of the combinatorial state space. Yet, SDVRPs are equally challenged by combinatorial action spaces. In essence, each decision point reduces to a vehicle routing problem given by Equation~(\ref{eq:static}). A holistic approach considers the service, routing, and assignment actions present in the VRP in one policy. 
%This yields an action space dimension that is the product of the number of requesting customers, the number of vehicles in the fleet, and the length of vehicles' routes. 
The resulting high-dimensional action space is impractical for solution approaches that approximate the value of state-action pairs as they typically enumerate all possible actions. Therefore, researchers have avoided searching the entire action space and restricted it, instead. That is, they have used RL to learn the values of only part of the action space and solved for another portion heuristically, as illustrated in the following example.\par

\cite{chen2019deep} face a same-day deliver problem with a heterogeneous fleet of drones and vehicles. A deep Q-learning policy learns whether to serve a delivery request by drone, vehicle, or reject it altogether. The assignment onto the fleet of vehicles or drones as well as the integration of deliveries into existing routes are performed by hand-crafted heuristics. In another example, \cite{joe2020deep} learn routing actions for a dynamic delivery problem. They assume that requests are already assigned to the fleet at the start of the day. Their policy combines Q-learning and simulated annealing to re-route vehicles when customers cancel or when new customers are assigned to a vehicle.\par

While action-space restrictions are often highly beneficial, in general, the resulting decision policy is coarse because the decomposed actions are dependant. There are few works on related dynamic problems that avoid such a brute-force approach. \cite{topaloglu2006dynamic} and \cite{chen2018approximate} combine linear or piece-wise linear value function approximations (VFAs) with powerful MIP solvers to search the action space more effectively. However, linear VFAs are known to poorly approximate state-action values for routing problems (see \cite{ulmer2020meso}). Therefore, we see an opportunity for combining MIP solvers with piecewise-linear functions that are capable of approximating the value function well. Most feed-forward neural networks rely on ReLu non-linearities which are, in fact, piece-wise linear. Therefore, we may think of trained ReLu neural networks as piece-wise linear functions which can be plugged into the MIP. In the past, ReLu constraints where modelled as big-M constraints which led to inefficient MIPs. Recently, \cite{anderson2020strong} and \cite{tsay2021partition} derived strong MIP formulations for trained ReLu neural networks. \cite{NEURIPS2020_06a9d51e} use their formulations and guide MIP solvers on static capacitated vehicle routing problems. However, to the best of our knowledge, a related work for a dynamic vehicle routing problem does not exist.

\subsubsection{Evaluate.}
Evaluating state-action pairs is difficult due to the combinatorial state space.
The space is so vast that most states are only observed once during millions of simulation runs and many state-action pairs are never explored. The high dimensionality makes naive tabulation approaches of state-action values impossible and encourages overfitting when carelessly employing DNNs as functional approximators. Thus, previous works on SDVRP heavily reduced the state space based on expert knowledge before learning the value of actions on the aggregated state space.
%\textcolor{green}{kommt hier nicht rein, aber ich könnte mir vorstellen, dass die schlupfvariablen im MIP coole feature sein könnten. nächste Masterarbeit, schreibs auf (:}
%Since these works use value-based RL approaches exclusively, and the fineness of the state space reduction depends on the approximation architecture, we give a short reminder on value-based RL methods.\par 

%\textcolor{red}{Perhaps we should delete this explanation of value-based methods? Jo, wir brauchen das detail hier nicht und haben an anfang der section jetzt schon ne kurze übersicht über RL.}
%Value-based approaches typically approximate the expected value $V^{\pi}(S_k, x_k)$ of choosing an action $x_k$ in state $S_k$ and following the current policy $X^\pi$ from there on. Sometimes, even learning state-values $V^{\pi}(S_k)$ is viable when paired with roll-out methods (see \cite{ulmer2019offline}). Approximating the value of states or state-action pairs are motivated by the same idea: A policy that always chooses the action that maximizes the optimal value function is an optimal policy. In practice, the optimal value function is unknown and we rely on a value function approximation (VFA), instead. Thus, our policy is implicitly defined by the VFA and we refine the VFA until we obtain a satisfactory, but generally not optimal, policy.\par
Value-based solution approaches differ in the approximation architecture and its update procedure. Many works on SDVRP rely on look-up tables (see \cite{schmid2012solving}, \cite{ulmer2018budgeting}, \cite{ulmer2019offline}) due to their ease of implementation and ability to approximate arbitrary value functions. Here, the biggest challenge is fine-tuning the granularity of the look-up table which is essentially a form of state space reduction. Some works dynamically reduce the state space by splitting interesting state-cells in the look-up table into a finer mesh as proposed by \cite{ulmer2018budgeting}. %While look-up tables perform well on routing problems, linear VFAs are better suited for knapsack problems (as shown in \cite{ulmer2020meso}). A prominent example of a knapsack-like problem in SDVRP is attended home delivery where a value function is estimated based on the remaining capacity of the fleet to solve a matching or pricing problem between customers and time-slots (see \cite{yang2017approximate}). 
Parametric VFAs are an alternative. Linear VFAs have the advantage that they can be plugged into an MIP formulation. However, when employing linear VFAs we assume that action values depend linearly on state information which is not the case for most SDVRPs. 
Other works hand-craft more general parametric VFAs based on expert knowledge and learn the parameters via value function iteration (see \cite{maxwell2010approximate} and \cite{al2020approximate}). Such a problem specific approach rarely translates to other problem settings.
Recently, some works on SDVRP started experimenting with neural networks to parameterize VFAs. There are multiple works that employ deep Q-networks to approximate state-action pair values (see \cite{chen2019deep} and \cite{chen2020same}). Such a framework has two advantages: 
\begin{enumerate}
    \item[a)] it is efficient as the action-values are obtained with only one forward pass through the Q-network,
    \item[b)] a well-chosen network architecture enables the effective usage of more features which is akin to a finer discretization.
\end{enumerate}
Advantage b) is the great promise that deep neural networks offer on the state space side of things.
However, larger feature spaces and corresponding larger networks come at the cost of longer training times which is still a limiting factor for real-world sized problem instances. Future work should propose mechanisms to train policies on small instances before transferring them to real-world instances.

\subsubsection{Key Takeaways}
A key challenge in solving SDVRPs is the combinatorial action space. Previous works strongly restricted the action space to choose actions via enumeration. In future, exact solvers could optimize over a piece-wise linear trained ReLu network that serves as VFA. Such a holistic approach promises to greatly improve solution quality.
A second challenge is the combinatorial state space. Past approaches strongly focused on coarse state space aggregation to enable value-based RL methods such as look-up tables. This came at the cost of solution quality. More recent approaches aim to capture more of the state information through the use of deep neural networks. Training these DNNs in a simulation environment is computationally expensive. Future solution methods need to include mechanisms to translate DNN policies from training instances to real-world sized applications.

\subsection{The CS Stream}
SDVRPs considered by the CS-community often resemble resource allocation problems. Typical examples include ride-sharing problems (see \cite{kullman2020dynamic}) and courier dispatching problems (see \cite{chen2019can}). In both problems, vehicles are assigned to customer requests or repositioned on a grid. No tentative routes must be planned and feasibility of actions can be verified directly. 
The lack of routes allows modelling the state space as a tensor of demand and vehicle positions over the grid. The action space entails the next grid tile to visit for each vehicle and can simply be enumerated. No state space reduction and action space restriction is needed and advanced RL methods can be used. One exception being \cite{balaji2019orl} who consider SDVRPs with time-windows and pickup-and-delivery constraints. However, they do not construct tentative routes.\par
In the following, we consider VRPs where enumeration of the action space is not feasible and present work that employs RL to \emph{search} the action space. Then, we discuss how CS-works \emph{evaluate} state-action pairs using full state information.

\subsubsection{Search.}
Most of the CS-community addresses SDVRPs with a small action space where no sophisticated search mechanism is needed. However, there is ML-work on more elaborate action spaces in the static VRP-literature (see \cite{bello2016neural}, \cite{deudon2018learning}, \cite{nazari2018reinforcement}, \cite{kool2018attention}, \cite{peng2019deep}, and \cite{sheng2020pointer}; for an overview see \cite{bai2021analytics})\footnote{We note that \cite{nazari2018reinforcement} and \cite{kool2018attention} also consider stochastic VRPs in their appendices.}. They model static vehicle routing problems with one vehicle as Markov decision processes. This way, an action is the next tentative stop to visit and not a whole tentative route. Then, a policy iteratively constructs routes without the need for enumerating the entire route-action space.
Such an approach could prove effective for solving VRP-subproblems in SDVRPs as they avoid action space restriction. Instead of using value-based RL methods, they employ policy-based methods that do not rely on value function approximation. In fact, in many SDVRPs the policy might just be the simpler function to approximate compared to the value function (see \cite[Chapter~13.1]{sutton2018reinforcement}). The reason can be found in SDVRP's high-dimensional state-action space and transition space which lead to high variance in rewards and, therefore, complex value functions. Yet, a strong policy can be very simple. 
However, a proof of effectiveness of policy-based methods for VRPs with more than one vehicle is required before they can be extended to SDVRPs. Furthermore, such an approach needs to be enhanced to address complex route constraints. An extra head in the policy network to assess the likelihood of meeting future route constraints is a possible remedy.

\subsubsection{Evaluate.}
\cite{chen2019can} and \cite{kullman2020dynamic}  learn state-action values with the help of well-chosen network architectures and update procedures. They rely on discretization of the service area to reduce the action space and choose actions via enumeration.\par
For the example of electric ride-sharing, \cite{kullman2020dynamic} split the state into system time, request information, and vehicle information. Notably, they construct their network architecture around the state space as opposed to reducing the state space in conformation to a given approximation architecture (see linear VFAs or look-up tables). %Specifically, they employ a deep dueling double Q-network with $n$-step temporal difference learning and prioritized experience replay.
%That is, the output of their Q-network is split into expected state value and action advantages (dueling), they use two independent Q-networks in their temporal difference update step to reduce bias in action value estimation (double), they incorporate the $n$ next discounted rewards in their update step ($n$-step temporal difference learning), and they sample state-action-reward tuples from a memory based on the sample's importance to the learning process (prioritized experience replay). 
%They embed the network in a multi-agent reinforcement learning framework that makes de-centralized action-value predictions for each vehicle. An attention mechanism helps the inner deep Q-network to make a centralized action for the fleet. \cite{kullman2020dynamic} show that this framework enables them to transfer policies trained on small instances to large-scale instances.\par
They embed the network in a multi-agent reinforcement learning framework and show that this framework enables them to transfer policies trained on small instances to large-scale instances.\par
\cite{chen2019can} use an actor-critic policy embedded in a multi-agent framework for the dynamic courier dispatching problem. They use purely de-centralized decision making but incentive cooperative behavior by shaping the reward function as a weighted sum of the courier's reward and the average reward of all other couriers. They show that their trained policy performs well out of distribution, i.e., when the dynamic of the test problem is significantly different to the training problem's dynamics.\par
We see an opportunity for future work in enhancing multi-agent reinforcement learning approaches with MIP formulations that model the complex interaction between independent agents to enable centralized decision-making.

\subsubsection{Key Takeaways.}
Complex SDVRPs with VRP-subproblems can be tackled by constructing tentative routes iteratively via a policy-based method. However, there is no example where such an approach proved effective for larger fleets. For simpler SDVRPs, where the entire state and action space can be considered directly, well-chosen Q-network architectures embedded in multi-agent reinforcement learning frameworks scale to larger instances and perform well out of distribution. In future, centralized decision-making for the independent agents could be achieved via MIP control.

\subsection{Summary}
We summarize the discussed works on RL for SDVRP in Tab.~\ref{tab:literature}. The column \enquote{Literature} cites the work. The column \enquote{Route Constraints} indicates whether the SDVRP comes with complex route constraints that require the construction of tentative routes.  The column \enquote{Full Action Space} denotes if the method searches or enumerates the entire action space. The column \enquote{Full State Space} indicates whether full state information is used in the solution method. We separate OR-work (top) and CS-work (bottom).\par
The table confirms our previous observations: There is no work in OR that considers a detailed action space for SDVRPs with complex route constraints. Further, there is no work in CS that addresses an SDVRP with complex route constraints.
\begin{table}[!t]
\caption{Classification of Literature Using RL for SDVRP}
\label{tab:literature}
\centering
\tiny 
\begin{tabular}{clccc}
\rowcolor[HTML]{C0C0C0} 
                     & Literature                              & \begin{tabular}[c]{@{}l@{}}Route\\ Constraints\end{tabular} & \begin{tabular}[c]{@{}l@{}}Full\\ Action Space\end{tabular} & \begin{tabular}[c]{@{}l@{}}Full\\ State Space\end{tabular} \\
                     & \cite{maxwell2010approximate}                                 &                                                             &                                                                     &                                                                 \\
                     & \cellcolor[HTML]{EFEFEF}\cite{schmid2012solving}          & \cellcolor[HTML]{EFEFEF}                                    & \cellcolor[HTML]{EFEFEF}                                            & \cellcolor[HTML]{EFEFEF}                                        \\
                     & \cite{chen2019deep}                               & \checkmark                                                   &                                                                     & \checkmark                                                       \\
                     & \cellcolor[HTML]{EFEFEF}\cite{ulmer2019offline}      & \cellcolor[HTML]{EFEFEF}\checkmark                           & \cellcolor[HTML]{EFEFEF}                                            & \cellcolor[HTML]{EFEFEF}                                        \\
                     & \cite{ulmer2020meso}                       & \checkmark                                                   &                                                                     &                                                                 \\
                     & \cellcolor[HTML]{EFEFEF}\cite{lei2020path}  & \cellcolor[HTML]{EFEFEF}\checkmark                           & \cellcolor[HTML]{EFEFEF}                                            & \cellcolor[HTML]{EFEFEF}                                        \\
                     & \cite{al2020approximate}                      &                                                             &                                                                     & \checkmark                                                       \\
                     & \cellcolor[HTML]{EFEFEF}\cite{chen2020same} & \cellcolor[HTML]{EFEFEF}\checkmark                           & \cellcolor[HTML]{EFEFEF}                                            & \cellcolor[HTML]{EFEFEF}\checkmark                               \\ 
 \multirow{-9}{*}{\rotatebox{90}{OR}} & \cite{joe2020deep}                         & \checkmark                                                   &                                                                     & \checkmark                                                       \\ \hline
                     & \cellcolor[HTML]{EFEFEF}\cite{nazari2018reinforcement}          & \cellcolor[HTML]{EFEFEF}                                    & \cellcolor[HTML]{EFEFEF}\checkmark                                   & \cellcolor[HTML]{EFEFEF}\checkmark                               \\
                     & \cite{kool2018attention}                                    &                                                             & \checkmark                                                           & \checkmark                                                       \\
                     & \cellcolor[HTML]{EFEFEF}\cite{chen2019can}        & \cellcolor[HTML]{EFEFEF}                                    & \cellcolor[HTML]{EFEFEF}\checkmark                                   & \cellcolor[HTML]{EFEFEF}\checkmark                               \\
 & \cite{balaji2019orl}                                &                                                             & \checkmark                                                           & \checkmark       \\

\multirow{-5}{*}{\rotatebox{90}{CS}} & \cellcolor[HTML]{EFEFEF}\cite{kullman2020dynamic}                                &    \cellcolor[HTML]{EFEFEF}  & \cellcolor[HTML]{EFEFEF} \checkmark                                                           & \cellcolor[HTML]{EFEFEF} \checkmark  
\end{tabular}
\end{table}

\section{Outlook \& Conclusion}\label{sec:outlook}

%\textcolor{red}{Ich würde hier noch mehr auf die MIP+evaluate challenge eingehen. Auch cool wäre ein central de-central ansatz: multiagent (kleine MIPs) mit globalerer steuerung. früher drauf hinweisen (bei Nick), dass agent-based action-space decomponiert und full search per agent zulässt. RL for routing action learning (state and future), SL for action feasibility}
%\textcolor{red}{summary raus und/oder auf die main insights reduzieren. We have identified...der nächster Absatz macht das eigentlich} Reinforcement learning has so far been used conservatively for the stochastic dynamic vehicle routing problem. Past works were challenged by combinatorial state and action space and resorted to drastic state aggregations and action space restrictions. Other works simplified the problem and searched the entire action space using full state information. In our work, we have provided a survey on work from both, the operations research and computer science community, to highlight current barriers and to point out future research directions.\par
Our analysis reveals many opportunities for future work. SDVRPs require powerful tools to \emph{search} the action space and \emph{evaluate} actions based on high-dimensional state information. OR- and CS-research provide the methodology to do either but not both. In future, a convergence of methodology promises holistic solutions. Concretely, we see potential in
\begin{itemize}
    \item Hybrid approaches that combine piece-wise linear neural network VFAs and strong COP solvers to \emph{search} the vast action space under \emph{evaluation} of future uncertainty. This requires strong MIP formulations and size-optimized networks to allow for real-time decision-making.
    \item \emph{Searching} for routes via policy-based methods that avoid the combinatorial action space by deciding iteratively on the next tentative stop to visit until a tentative route is constructed. An extra head in the neural network could \emph{evaluate} the long-term feasibility of adding a tentative stop to the route to avoid future violation of service promises. However, policy-based approaches are completely unexplored for more than one vehicle or dynamic routing. Thus, methods must not only learn to find the route that minimizes current costs but also include some evaluation of the future. %Potentially, a multi-agent reinforcement approach is required to extend to larger fleets.
    \item Multi-agent reinforcement learning approaches that enhance de-centralized %agent's state 
    \emph{evaluation} of all of the individual agent's actions with a global MIP \emph{searching} the joint action space. Multi-agent approaches facilitate scaling to larger instances and accommodate day-to-day changes in fleet size. However, we lose information when controlling multiple agents as we effectively decompose the action space. 
    %\item \emph{Evaluating} actions with the help of deep neural network architectures that capture complex relations between resources, demand, and environment. This comes at the drawback that it potentially limits \emph{searching} the action space.
\end{itemize} 

\bibliographystyle{named}
\bibliography{main}

\end{document}